\documentclass[sigconf]{acmart}

\AtBeginDocument{%
  }

\setcopyright{acmlicensed}
\copyrightyear{2025}
\acmYear{2025}
\setcopyright{acmlicensed}
\acmConference[CIKM '25] {Proceedings of the 34th ACM International Conference on Information and Knowledge Management}{ November 10--14, 2025}{Seoul, Republic of Korea.}
\acmBooktitle{Proceedings of the 34th ACM International Conference on Information and Knowledge Management (CIKM '25), November 10--14, 2025, Seoul, Republic of Korea}
\acmISBN{979-8-4007-2040-6/2025/11}
\acmDOI{10.1145/3746252.3761614}
\acmSubmissionID{1328}

\usepackage{microtype}
\usepackage{hyperref}
\usepackage{url}
\usepackage{booktabs}

\usepackage{graphicx}

\usepackage{algorithm}
\usepackage{algpseudocode}

\usepackage{wrapfig}
\usepackage{graphicx}
\usepackage{textcomp}
\usepackage{xcolor}
\usepackage[normalem]{ulem}
\useunder{\uline}{\ulined}{}%
\DeclareUrlCommand{\bulurl}{}

\usepackage{booktabs,enumitem} 
\usepackage[nameinlink,capitalise]{cleveref}

\usepackage{pifont} 

\newcommand{\cmark}{\ding{51}} 
\newcommand{\xmark}{\ding{55}} 

\newcommand{\HeartCapEmoji}{\raisebox{-0.45em}{\includegraphics[height=1.4em,trim=0 .4em 0 0]{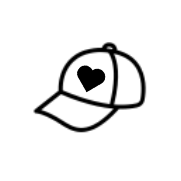}}}

\begin{document}

\title{S2Cap: A Benchmark and a Baseline for Singing Style Captioning}

\author{Hyunjong Ok}
\affiliation{%
  \institution{Pohang University of Science and Technology}
  \city{Pohang}
  \country{South Korea}
}
\email{hyunjong.ok@postech.ac.kr}

\author{Jaeho Lee}
\affiliation{%
  \institution{Pohang University of Science and Technology}
  \city{Pohang}
  \country{South Korea}
}
\email{jaeho.lee@postech.ac.kr}

\renewcommand{\shortauthors}{Ok and Lee}

\begin{abstract}
Singing voices contain much richer information than common voices, including varied vocal and acoustic properties. However, current open-source audio-text datasets for singing voices capture only a narrow range of attributes and lack acoustic features, leading to limited utility towards downstream tasks, such as style captioning. To fill this gap, we formally define the singing style captioning task and present S2Cap, a dataset of singing voices with detailed descriptions covering diverse vocal, acoustic, and demographic characteristics. Using this dataset, we develop an efficient and straightforward baseline algorithm for singing style captioning. The dataset is available at \bulurl{https://zenodo.org/records/15673764}.

\end{abstract}

\begin{CCSXML}
<ccs2012>
   <concept>
       <concept_id>10010147.10010178</concept_id>
       <concept_desc>Computing methodologies~Artificial intelligence</concept_desc>
       <concept_significance>500</concept_significance>
       </concept>
 </ccs2012>
\end{CCSXML}

\ccsdesc[500]{Computing methodologies~Artificial intelligence}

\keywords{Singing style captioning, Dataset pipeline, Audio-to-text model}

\maketitle

\section{Introduction}

Following the recent progress in text-to-speech modeling, the task of \textit{speaking style captioning} has received a great deal of attention \citep{yamauchi2024stylecap, ando2024factor, zhu2024unistyle}. Here, the goal is to generate a text prompt that describes para-/non-linguistic characteristics of the speaker from the given audio clip, such as pitch, volume, or gender. The extracted information can contribute greatly to advancing the state-of-the-art of style-conditioned speech synthesis by providing a useful basis for evaluating and labeling the speech data \citep{wang2018style,guo2023prompttts, yang2024instructtts, leng2024prompttts}.

How much information can speaking style captioning models capture from the \textit{singing voices}? Singing voices contain rich musical characteristics, such as timbre, tempo, or musical genre, providing valuable information for the synthesis and conversion of singing voices \citep{luo2020singing, zhang2024stylesinger}. However, existing speaking style captioning models have been trained to extract only non-musical characteristics from the audio and thus may fall suboptimal for such a purpose. In fact, there is even a lack of appropriate benchmarks to evaluate the performance of such models. 
Although some datasets provide text attributes or prompts paired with the singing voices \citep{wang2022opencpop,huang2021multi,wang2024prompt}, they are limited in the scale of the dataset, the number of attributes, or the diversity of singers (see \Cref{table:compare_data}).

To fill this gap, we formally introduce the task of \textit{singing style captioning} and take the first step toward solving this task. We first present S2Cap (\underline{S}inging \underline{S}tyle \underline{Cap}tioning), a singing voice dataset labeled with ten vocal, musical, and demographic attributes; this is much larger than prior work---\textit{e.g.}, \citet{wang2024prompt} with only three attributes---enabling the trained model to understand detailed and diverse styles of the singing voices.

\begin{table}[t]
\centering
\vspace{0.25cm}
\caption{A comparison of the constructed S2Cap dataset with related singing voice datasets.}
\resizebox{0.9\columnwidth}{!}{
\begin{tabular}{lccccc}
\toprule
\textbf{Dataset} & \textbf{\# Singers} & \textbf{Duration (h)} & \textbf{Style Caption} & \textbf{\# Attributes} \\ \midrule
NUS-48E~\cite{6694316}               & 12 & 1.9 & \xmark & -  \\
Opencpop~\cite{wang2022opencpop}     & 1 & 5.3 & \xmark & -  \\ 
OpenSinger~\cite{huang2021multi}     & 66 & 50.0 & \xmark & -  \\ 
M4Singer~\cite{NEURIPS2022_2de60892} & 20 & 29.8  & \xmark & -  \\ 
Prompt-Singer~\cite{wang2024prompt}  & 758 & 306.9 & \cmark & 3  \\ \midrule
\textbf{S2Cap (Ours)}                & 2,376 & 262.9 & \cmark & 10 \\ \bottomrule
\end{tabular}
}
\label{table:compare_data}
\vspace{-0.9cm}
\end{table}

Building upon the S2Cap dataset, we provide comprehensive baselines by evaluating combinations of diverse audio encoders and text decoders alongside state-of-the-art in related tasks: audio captioning and music captioning. Additionally, we propose a novel training objective designed to enhance the model's focus on vocal information, utilizing demixed vocal audio as auxiliary supervision. This offers a path for future work to achieve better performance.

\begin{figure*}[t]
    \centering
    \includegraphics[width=0.78\linewidth]{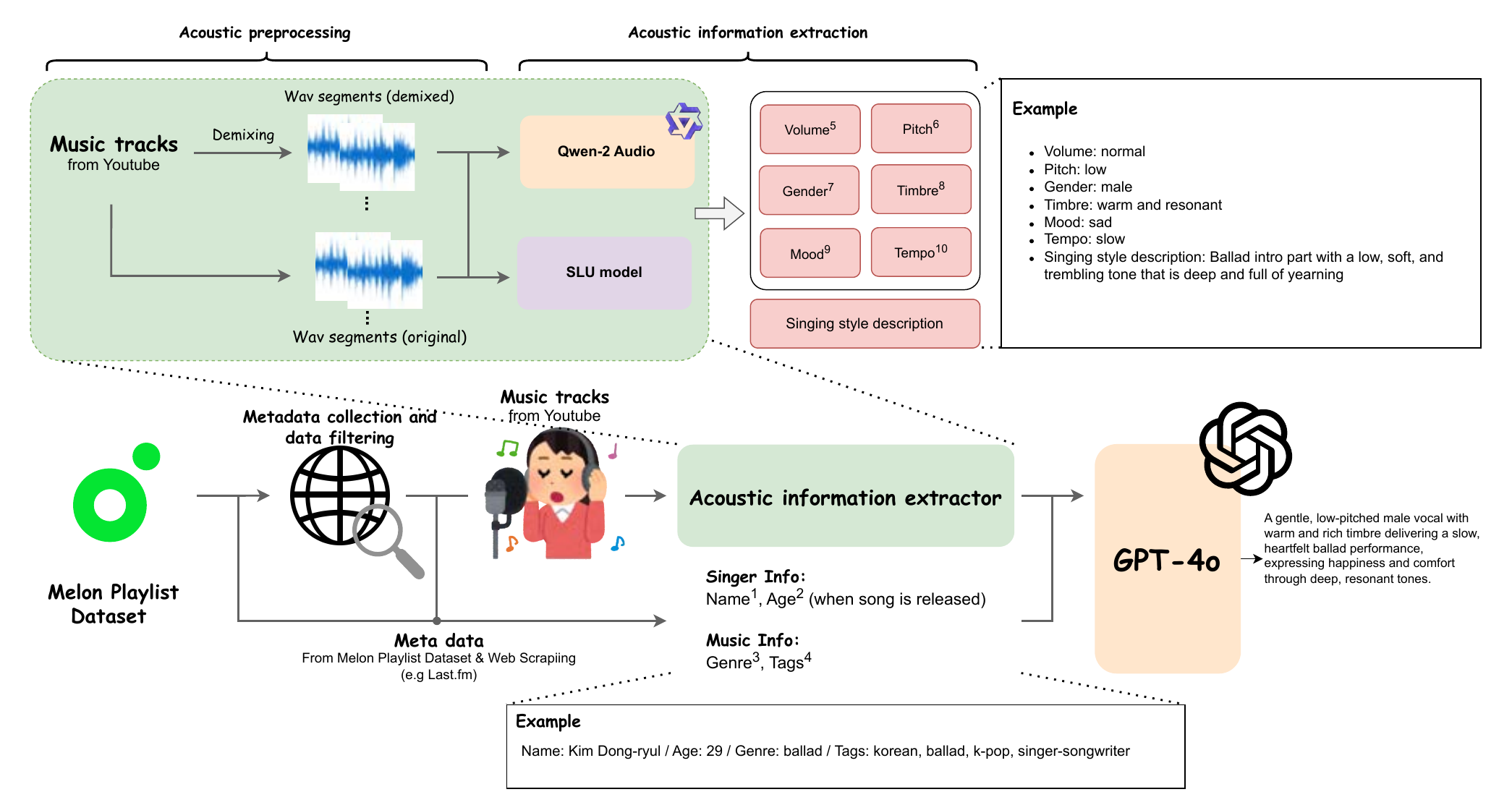}
    \caption{An illustration of the data generation pipeline of the S2Cap. We start from a base playlist dataset and collect metadata via web scraping. Then, we partition the audio tracks into multiple segments and extract acoustic information from each segment; through these steps, we collect a total of ten attributes (name, age, genre, tag, volume, pitch, gender, timbre, mood, tempo; marked with a subscript) and the singing style description. Finally, we summarize the collected attributes into a single textual prompt using GPT-4o.} \label{fig:datapipline_overview}
\end{figure*}

\section{Related work}

\paragraph{Captioning tasks and datasets.}
Captioning task aims to generate descriptive texts corresponding to input from diverse modalities. 

Various datasets in the visual domain facilitate research for visual captioning. For image captioning, datasets include Flickr30k \citep{young-etal-2014-image}, MS COCO \citep{chen2015microsoft}, and Conceptual Captions \citep{sharma-etal-2018-conceptual}, each offering diverse annotations for image descriptions. In video captioning, datasets such as MSVD \citep{chen2011collecting}, MSR-VTT \citep{7780940}, and LSMDC \citep{rohrbach2017movie} support research of textual descriptions from sequential visual content.

In the auditory domain, datasets such as AudioCaps \citep{kim2019audiocaps} and Clotho \citep{drossos2020clotho} have been used for general audio captioning tasks. Recently, there has been growing interest in speech style captioning, which involves describing characteristics of human speech (e.g., gender, age group, vocal range) using textual descriptions. PromptTTS \citep{guo2023prompttts} and InstructTTS \citep{yang2024instructtts}, for instance, construct dedicated datasets to learn rich speaker-style representations and employ Transformer-based architectures to integrate these representations into synthesized speech. PromptVC \citep{yao2024promptvc} focuses on text-prompted voice conversion, and PromptSpeaker \citep{zhang2023promptspeaker} aims to generate speaker embeddings from text prompts. StyleCap \citep{yamauchi2024stylecap} initiates speech style captioning to get style descriptions from speech. 

Building upon this research, we introduce a new task, singing style captioning, which extends captioning beyond generic speech to the domain of singing with a fine-grained dataset S2Cap that contains various attributes and abundantly explains the singing style through text descriptions.

\paragraph{Music source separation.}
Music source separation (MSS), also known as demixing, aims to decompose a mixed music audio signal into its constituent sources, such as vocals, drums, and other accompaniment. Recent advances in deep learning have significantly improved MSS performance, leveraging architectures such as CNN, RNN, and transformer-based networks. Band-split RNN \citep{luo2023music} and HT Demucs \citep{rouard2023hybrid} have emerged as state-of-the-art models. Band-split RNN employs an RNN-based approach processing different frequency bands separately, enabling effective harmonic modeling and improving MSS accuracy. HT Demucs utilizes transformer architecture to enhance separation performance, capturing local and global dependencies within audio signals.

In our work, we utilize HT Demucs to extract vocal WAV files, which serve as input for our data generation and method.

\section{\HeartCapEmoji S2Cap}
\label{subsec:method_1}

We now introduce S2Cap, a singing style captioning dataset. The S2Cap consists of 12,105 music tracks with 71,215 textual captions that describe singing styles; each track is partitioned into multiple segments, on which we put separate captions. Along with the captions, each segment has annotations on ten attributes (name, age, genre, tags, volume, pitch, gender, timbre, mood, and tempo). The detailed statistics are given in \cref{table:data-stat}.

To avoid license issues, we do not directly share the WAV files for the audio tracks. Instead, we provide the code and URLs to download the wav files. 

\begin{table}[t]
\centering
%\vspace{0.25cm}
\caption{Basic statistics of the S2Cap dataset.}
\resizebox{0.9\columnwidth}{!}{
\begin{tabular}{lccccc}
\toprule
\textbf{Splits} & \textbf{Tracks}& \textbf{Captions} & \textbf{Words/Caption} & \textbf{Total tokens} & \textbf{Duration (h)} \\ \midrule
Train & 8,395 & 49,325 & 23.28 & 1,148,142 & 181.86 \\ 
Dev   & 1,232 & 7,339  & 23.23 & 170,500  & 27.22 \\ 
Test  & 2,478 & 14,551 & 23.30 & 339,089  & 53.81 \\ \midrule
Total & 12,105 & 71,215 & 23.28 & 1,657,731 & 262.88 \\ \bottomrule
\end{tabular}
}
\label{table:data-stat}
\end{table}

\subsection{Construction Pipeline}
For the sake of scale, the S2Cap dataset has been constructed by processing an existing musical dataset with an LLM-based pipeline to generate corresponding textual captions (see \cref{fig:datapipline_overview}). As the base dataset, we have used the Melon playlist dataset, a public dataset containing Mel-spectrogram and textual metadata with over 600,000 tracks \citep{ferraro2021melon}. Although it originates from a Korean music streaming service, ``Melon,'' it covers a wide range of international music across diverse genres and artists, beyond Korean songs. Each track in the dataset has been processed as follows.

\paragraph*{Metadata collection and data filtering.} First, we collect textual metadata for each track by scraping from various web sources, \textit{e.g.}, Last.fm tags. 
Then, we preprocess the audio track and the metadata as follows. (1) Since the Mel-spectrogram in the base dataset is of low resolution to avoid licensing issues, we collect higher-quality audio tracks from YouTube. (2) We filter out the audio files with missing or mismatched metadata entries; we also filtered out the songs from singers born before 1970, as their songs tend to be missing on YouTube.

\paragraph{Acoustic preprocessing.} Before extracting the acoustic features, we extract two versions of the audio track: the original version and the vocal-only version. The vocal-only version is useful for capturing the style of the vocalist separately from the instrumental parts, and the original version comes in handy in capturing the overall mood. The vocal-only version is prepared using a demixing model, namely the HT Demucs \citep{rouard2023hybrid}. For handling the case of multiple singers with different styles, the demixed tracks are further processed with a speaker diarization model\footnote{\url{https://huggingface.co/pyannote/speaker-diarization-3.1}}; the audio clips are partitioned into 30 seconds-long segments for effective processing, following the prior works \citep{agostinelli2023musiclm,doh2023lp,kim2019audiocaps}. Any segment shorter than 5 seconds is discarded. We also apply the same segmentation to the original version of the audio track. 

\paragraph*{Acoustic information extraction.} Next, we extract acoustic attributes (volume, pitch, gender, timbre, mood, tempo) and style description prompts from each audio segment. This is done with two pretrained models: Qwen-2 Audio \citep{chu2024qwen2}, and a spoken language understanding (SLU) model.\footnote{\url{https://github.com/JeremyCCHsu/Python-Wrapper-for-World-Vocoder}} Qwen-2 Audio is used to generate annotations on four attributes (gender, timbre, mood, tempo) and the singing style description text; gender and timbre are extracted from the vocal-only version of the audio, and others are generated using the original version. SLU is used to generate volume and pitch annotations. Here, both attributes are classified into three categories---``low,'' ``medium,'' and ``high''---where the volume is determined based on the root-mean square of the amplitude and the pitch is determined based on the average $F_0$.

\paragraph*{Prompt generation.} Finally, the extracted attributes and singing style description are summarized into a single textual prompt (per segment), with GPT-4o\footnote{in particular, the GPT-4o-2024-08-06} \citep{hurst2024gpt}. 

\paragraph{Data splitting.}
We have partitioned the S2Cap dataset into training/development/test sets in the 70\%/10\%/20\% ratio. We split the dataset so that no artist appears in multiple subsets. Additionally, we have balanced the distribution of six acoustic attributes to preserve the statistical consistency across splits.

\paragraph{Omitted details.} More detailed information and code are available at \bulurl{https://github.com/HJ-Ok/S2cap}.

\section{Experiments}

\paragraph{Baselines.} 
We establish comprehensive baselines for our task. 
We evaluate transformer-based architectures, aligning with our proposed methodology's framework. We conduct an extensive ablation study across various audio encoder and text decoder combinations. We evaluate four pretrained models for audio encoding: AST, MERT \citep{li2024mert}, Wav2vec 2.0, and HuBERT \citep{hsu2021hubert}. These are systematically paired with two decoder variants: GPT-2 and BART-base (w/ decoder part only) \citep{raffel2020exploring}, exploring eight encoder-decoder configurations.
Also, we include two specialized models, Prefix-AAC \citep{kim2023prefix} and LP-MusicCaps \citep{doh2023lp}. They represent the state-of-the-art in related tasks: audio captioning and music captioning. 
All models are finetuned for $20$ epochs with a batch size of $32$, accumulation steps of $2$, weight decay of $2$, and learning rate of $2 \times 10^{-5}$. We use beam search with beam size $5$ during inference.

\paragraph{Evaluation.}
To evaluate our proposed methods, we employ metrics that are widely used in captioning tasks such as BLEU \citep{10.3115/1073083.1073135}, METEOR \citep{banerjee-lavie-2005-meteor}, ROUGE-L \citep{lin-2004-rouge}, CIDEr \citep{vedantam2015cider}, SPICE \citep{anderson2016spice}, and SPIDEr \citep{liu2017improved}. BLEU is a modified n-gram precision metric incorporating a brevity penalty, while ROUGE-L calculates an F-measure based on the longest common subsequence. METEOR enhances the evaluation by considering several factors like stem- and word-level overlap and synonymy. CIDEr employs a geometric mean of n-gram and cosine similarity scores. SPICE focuses on semantic content by parsing scene graphs from captions, and SPIDEr is the average score of SPICE and CIDEr. While the above metrics are valuable for assessing captioning systems, inspired by recent findings \citep{zhou2022can, 10096526}, they have limits in capturing the semantic meaning in generated captions. So we supplement our evaluation with Sentence-BERT \citep{reimers-gurevych-2019-sentence}, metrics tailored for improved semantic relatedness, which produces embeddings for calculating sentence-level similarity.

\paragraph{Additional demixing supervision method.}
To enhance the model's ability to effectively represent singing voices against background music, we introduce a novel fine-tuning strategy that incorporates a demixing supervision loss. This approach regularizes the audio encoder to focus on vocal signals. The training is done with a mixture of two loss functions, which we describe below.

Our training objective combines cross-entropy loss for caption generation with our novel demixing supervision loss. The cross-entropy loss employs teacher-forcing strategy, where the model receives the ground-truth token from the previous step as input:
\begin{equation}
    \mathcal{L_{\text{CE}}} = -\sum_{n=1}^{N} \log P(y_n | y_1, \ldots, y_{n-1}, E_{\text{audio}}(X))
\end{equation}
where $P(\cdot|\cdot)$ denotes the output probability of the text decoder and $Y = [y_1,y_2,\ldots,y_N]$ denotes the text prompt paired with audio $X$. The demixing supervision loss encourages the audio encoder to extract similar features from both original tracks and their vocal-only counterparts by minimizing the KL divergence:
\begin{equation} 
\mathcal{L_{\text{demix}}} = D_{\text{KL}}\left(E_{\text{audio}}(X_\text{voc}) \| E_{\text{audio}}(X)\right), 
\end{equation}
where $D_{\text{KL}}(\cdot\|\cdot)$ denotes the KL divergence between two embedding.

The final loss is the mixture of the cross-entropy and the demixing supervision:
\begin{equation}
\mathcal{L_{\text{final}}} = \mathcal{L_{\text{CE}}} + \lambda \cdot \mathcal{L_{\text{demix}}},
\end{equation}
where $\lambda$ is the weight hyperparameter. 
%We have simply used $\lambda=1$ in all experiments.

\begin{table}[t]
\footnotesize
\centering
    \vspace{0.25cm}

    \caption{Experiment results on the S2Cap test set of our method and baselines. For all metrics, the higher, the better.}\vspace{1mm}
    \resizebox{0.94\linewidth}{!}{
    \begin{tabular}{lccccccc}
    \toprule
    Methods & BLEU$_4$&METEOR&ROUGE-$L$&CIDEr&SPICE&SPIDEr&Sentence-BERT \\
    \midrule 
        \multicolumn{8}{c}{w/ GPT2}\\
    \midrule 
    AST-GPT2  & 26.56 & 30.68 & 52.64 & 100.73 & 44.05 & 72.39 & 84.18  \\
    MERT-GPT2  & 26.99 & 31.05 & 53.11 & 101.75 & 45.19 & 73.47 & 85.51  \\
    Wav2vec 2.0-GPT2  & 25.99 & 30.43 & 52.41 & 94.37 & 44.22 & 69.30 & 85.01  \\
    HuBERT-GPT2  & 25.80 & 30.56 & 52.45 & 95.41 & 44.62 & 70.01 & 85.28  \\

    \midrule 
        \multicolumn{8}{c}{w/ BART}\\
    \midrule 
    AST-BART  & 26.47 & 30.38 & 52.50 & 99.42 & 43.98 & 71.70 & 84.31  \\
    MERT-BART  & 26.49 & 30.43 & 52.50 & 99.95 & 43.92 & 71.93 & 84.04  \\
    Wav2vec 2.0-BART  & 25.51 & 30.01 & 51.97 & 93.30 & 43.81 & 68.56 & 84.53  \\
    HuBERT-BART  & 26.18 & 30.27 & 52.60 & 96.45 & 44.34 & 70.39 & 84.76  \\

    \midrule 
        \multicolumn{8}{c}{w/ T5}\\
    \midrule 
    AST-T5  & 29.12 & 31.50 & 56.25 & 105.19 & 43.24 & 74.21 & 85.32  \\
    MERT-T5  & 21.78 & 25.30 & 49.88 & 58.99 & 34.16 & 46.58 & 81.62  \\
    Wav2vec 2.0-T5  & 28.83 & 31.12 & 56.37 & 102.77 & 44.57 & 73.67 & 86.25 \\
    HuBERT-T5  & 27.64 & 30.78 & 55.67 & 98.07 & 43.21 & 70.64 & 85.52  \\

    \midrule 
        \multicolumn{8}{c}{SOTA of related works}\\
    \midrule 
    Prefix-AAC  & 29.47 & 31.64 & 56.84 & 104.10 & 44.87 & 74.48 & 86.38  \\
    + w/ Demixing supervision & 29.70 & 31.74 & 56.97 & 105.70 & 44.89 & 75.29 & 86.45 \\

    LP-MusicCaps  & 28.33 & 31.06 & 55.60 & 102.92 & 42.61 & 72.77 & 84.91  \\

    \bottomrule
    \end{tabular}
     }
\label{table:baselines}
\end{table}

\begin{figure*}[t]
    \centering
    \includegraphics[width=0.7\linewidth]{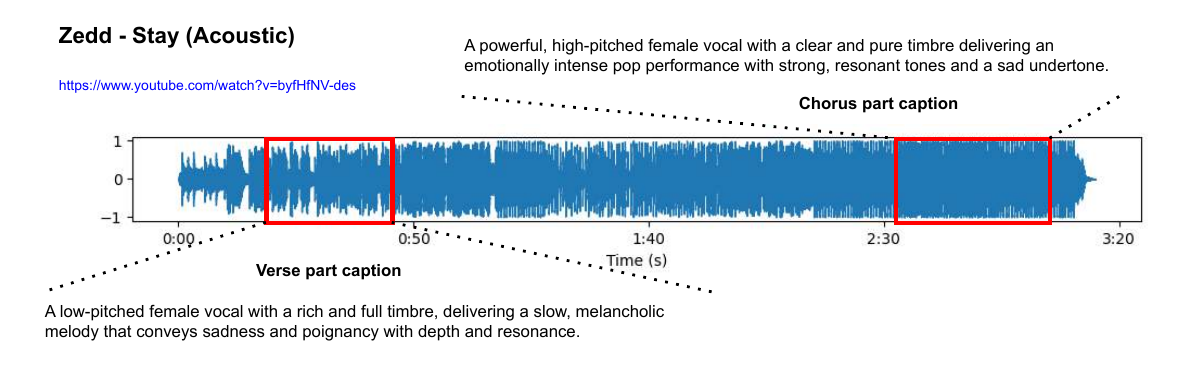}
    \caption{An example from S2Cap. The emotional intensity of the music gradually escalates from the first to the latter half, effectively captured in the generated caption.
    }
    
    \label{fig:case_study}
\end{figure*}

\section{Results}

\subsection{Experiment results}

\paragraph{Experiments in various baselines.}
We conducted experiments using various encoder-decoder combinations and state-of-the-art (SOTA) models from related works, which are shown in \cref{table:baselines}. Additionally, we applied demixing supervision to the best-performing model Prefix-AAC from our baseline experiments, demonstrating improved performance with demixing supervision.

\subsection{Data quality assessment}

\begin{figure}[t]
    \centering
    \includegraphics[width=0.75\columnwidth]{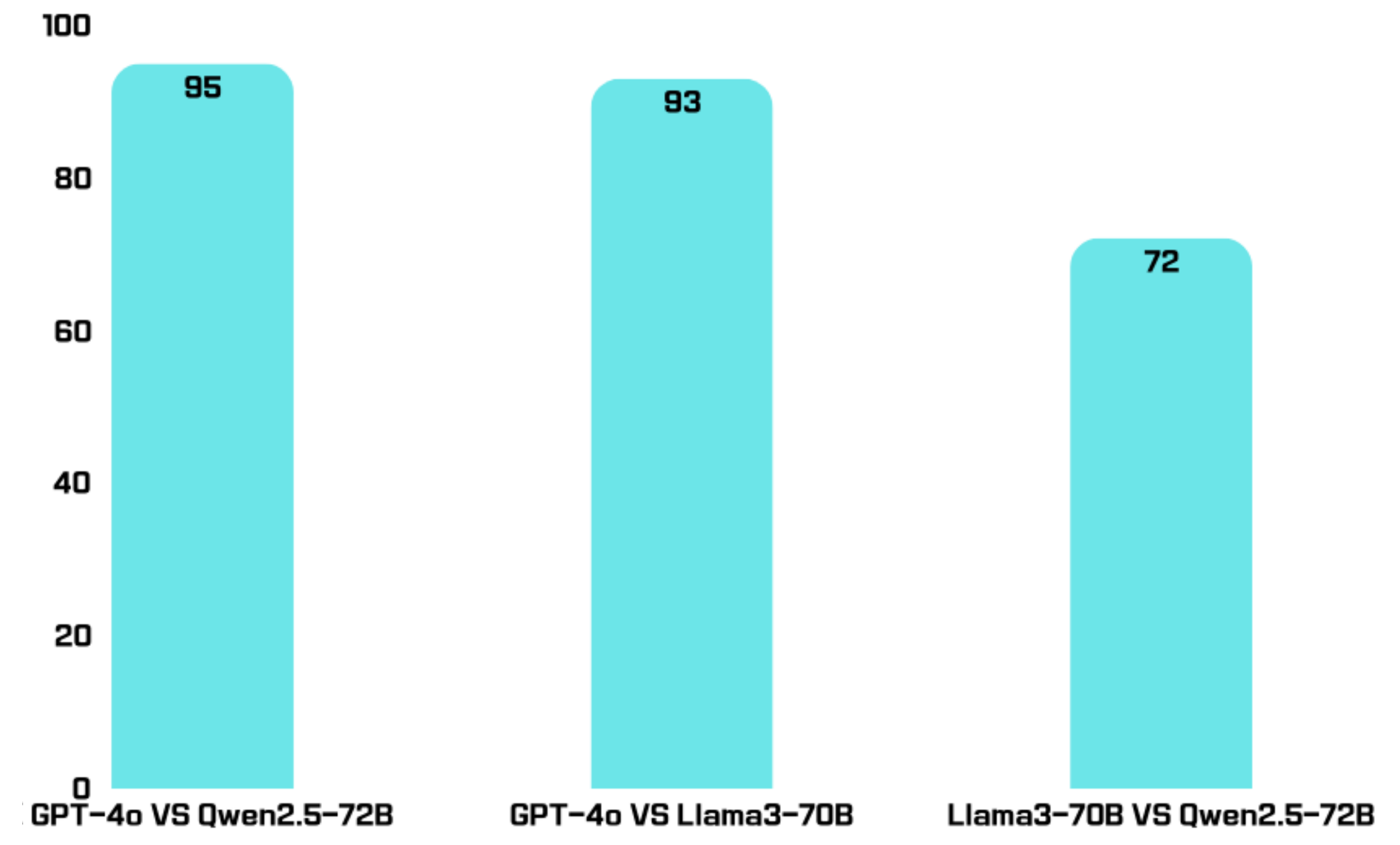}
    \caption{Human evaluation results comparing generated captions. The numbers in the bar plot indicate the win rate of the model on the left.
    }
    
    \label{fig:human_eval}
\end{figure}

\paragraph*{Human evaluation.}
To assess the quality of our dataset, we conducted a human evaluation study by comparing captions generated by GPT-4o, Qwen2.5-72B-Instruct \citep{yang2024qwen2}, and Llama3.3-70B-Instruct \citep{dubey2024llama} models. Three human annotators were given 200 sampled captions with audio and asked to determine which model produced the best outputs. As shown in \cref{fig:human_eval}, GPT-4o consistently outperformed other models, demonstrating superior caption quality.

In addition to comparative evaluation, we further analyzed GPT-4o-generated caption quality by assessing consistency and fluency criteria, scored out of 5 and 3, respectively. Consistency was evaluated based on factual alignment with music audio tracks, while fluency was measured in terms of grammar, spelling, punctuation, word choice, and sentence structure. Three human annotators were given the same 200 sampled captions with audio and asked to evaluate caption quality. As shown in \cref{tab:human_evaluation}, GPT-4o achieved an average consistency score of 4.94 and fluency score of 2.97, confirming the high quality of its generated captions.

To check the objectivity, we have conducted a human evaluation of the timbre generated by Qwen-2 Audio, where the subjective terms come from. Specifically, 20 annotators judged whether the generated timbre appropriately matched one of the 20 well-known singers. Timbre accuracy is 0.75, indicating a strong alignment between human perception and Qwen-2 Audio. Recent studies report human-LLM alignment in evaluations typically about 70\%~\citep{kolchinski2019approximating, yang2024air}, our result demonstrates notably high agreement. 

\begin{table}[t]
\caption{The result of human evaluation on the quality of captions generated by S2Cap.
} 
\centering
\resizebox{0.65 \columnwidth}{!}{%
\begin{tabular}{lccc}
\toprule
 Methods & \multicolumn{1}{|c}{Consistency} & Fluency & \multicolumn{1}{|c}{ Timbre Acc.}    \\ \midrule
Human eval & 4.94 & 2.97 & 0.75 \\
\bottomrule
\end{tabular}%
}
\label{tab:human_evaluation}
\end{table} 

\paragraph*{Captioning examples.}

We show an example of our S2Cap dataset, as illustrated in \cref{fig:case_study}. 
These captions effectively capture various attributes of singing, demonstrating high-quality caption generation. In the given example song, the verse part features a soft and emotional vocal, whereas the chorus gradually intensifies, culminating in an explosive emotional peak. Our dataset successfully reflects these dynamic shifts in the same song.

\section{Conclusion}
We propose a novel task, singing style captioning, which aims to generate textual prompts describing the vocal characteristics of singers from given song inputs. For this task, we developed S2Cap, a comprehensive dataset reflecting diverse vocal attributes, and established a robust baseline method that effectively captures singing voice characteristics. These contributions provide a solid foundation for future research in this emerging field.

\section*{Usage of Generative AI}
We use for manuscript refinement and code optimization. Additionally, LLMs are utilized into our data generation pipeline.

\begin{acks}
This work was supported by the National Research Foundation of Korea (NRF) grant funded by the Korea government (MSIT) (No. RS-2024-00453301), and by the Institute of Information \& communications Technology Planning \& Evaluation (IITP) grant funded by the Korea government (MSIT) (No.RS-2019-II191906, Artificial Intelligence Graduate School Program(POSTECH)).
\end{acks}

\bibliographystyle{ACM-Reference-Format}
\bibliography{references}


\begin{thebibliography}{48}

%%% the Web address.

\ifx \showCODEN    \undefined \def \showCODEN     #1{\unskip}     \fi
\ifx \showISBNx    \undefined \def \showISBNx     #1{\unskip}     \fi
\ifx \showISBNxiii \undefined \def \showISBNxiii  #1{\unskip}     \fi
\ifx \showISSN     \undefined \def \showISSN      #1{\unskip}     \fi
\ifx \showLCCN     \undefined \def \showLCCN      #1{\unskip}     \fi
\ifx \shownote     \undefined \def \shownote      #1{#1}          \fi
\ifx \showarticletitle \undefined \def \showarticletitle #1{#1}   \fi
\ifx \showURL      \undefined \def \showURL       {\relax}        \fi
% invisible to TeX
\providecommand\bibfield[2]{#2}
\providecommand\bibinfo[2]{#2}
\providecommand\natexlab[1]{#1}
\providecommand\showeprint[2][]{arXiv:#2}

\bibitem[Agostinelli et~al\mbox{.}(2023)]%
        {agostinelli2023musiclm}
\bibfield{author}{\bibinfo{person}{A. Agostinelli}, \bibinfo{person}{T.~I Denk}, \bibinfo{person}{Z. Borsos}, \bibinfo{person}{J. Engel}, \bibinfo{person}{M. Verzetti}, \bibinfo{person}{A. Caillon}, \bibinfo{person}{Q. Huang}, \bibinfo{person}{A. Jansen}, \bibinfo{person}{A. Roberts}, \bibinfo{person}{M. Tagliasacchi}, {et~al\mbox{.}}} \bibinfo{year}{2023}\natexlab{}.
\newblock \showarticletitle{Musiclm: Generating music from text}.
\newblock \bibinfo{journal}{\emph{arXiv preprint arXiv:2301.11325}} (\bibinfo{year}{2023}).
\newblock


\bibitem[Anderson et~al\mbox{.}(2016)]%
        {anderson2016spice}
\bibfield{author}{\bibinfo{person}{P. Anderson}, \bibinfo{person}{B. Fernando}, \bibinfo{person}{M. Johnson}, {and} \bibinfo{person}{S. Gould}.} \bibinfo{year}{2016}\natexlab{}.
\newblock \showarticletitle{Spice: Semantic propositional image caption evaluation}. In \bibinfo{booktitle}{\emph{ECCV}}.
\newblock


\bibitem[Ando et~al\mbox{.}(2024)]%
        {ando2024factor}
\bibfield{author}{\bibinfo{person}{A. Ando}, \bibinfo{person}{T. Moriya}, \bibinfo{person}{S. Horiguchi}, {and} \bibinfo{person}{R. Masumura}.} \bibinfo{year}{2024}\natexlab{}.
\newblock \showarticletitle{Factor-Conditioned Speaking-Style Captioning}.
\newblock \bibinfo{journal}{\emph{arXiv preprint arXiv:2406.18910}} (\bibinfo{year}{2024}).
\newblock


\bibitem[Banerjee and Lavie(2005)]%
        {banerjee-lavie-2005-meteor}
\bibfield{author}{\bibinfo{person}{S. Banerjee} {and} \bibinfo{person}{A. Lavie}.} \bibinfo{year}{2005}\natexlab{}.
\newblock \showarticletitle{{METEOR}: An Automatic Metric for {MT} Evaluation with Improved Correlation with Human Judgments}. In \bibinfo{booktitle}{\emph{{ACL} Workshop}}.
\newblock


\bibitem[Bhosale et~al\mbox{.}(2023)]%
        {10096526}
\bibfield{author}{\bibinfo{person}{S. Bhosale}, \bibinfo{person}{R. Chakraborty}, {and} \bibinfo{person}{S. Kopparapu}.} \bibinfo{year}{2023}\natexlab{}.
\newblock \showarticletitle{A Novel Metric For Evaluating Audio Caption Similarity}. In \bibinfo{booktitle}{\emph{ICASSP}}.
\newblock
\href{https://doi.org/10.1109/ICASSP49357.2023.10096526}{doi:\nolinkurl{10.1109/ICASSP49357.2023.10096526}}


\bibitem[Chen and Dolan(2011)]%
        {chen2011collecting}
\bibfield{author}{\bibinfo{person}{D. Chen} {and} \bibinfo{person}{W.~B Dolan}.} \bibinfo{year}{2011}\natexlab{}.
\newblock \showarticletitle{Collecting highly parallel data for paraphrase evaluation}. In \bibinfo{booktitle}{\emph{ACL}}.
\newblock


\bibitem[Chen et~al\mbox{.}(2015)]%
        {chen2015microsoft}
\bibfield{author}{\bibinfo{person}{X. Chen}, \bibinfo{person}{H. Fang}, \bibinfo{person}{T.-Y. Lin}, \bibinfo{person}{R. Vedantam}, \bibinfo{person}{S. Gupta}, \bibinfo{person}{P. Doll{\'a}r}, {and} \bibinfo{person}{C~L. Zitnick}.} \bibinfo{year}{2015}\natexlab{}.
\newblock \showarticletitle{Microsoft coco captions: Data collection and evaluation server}.
\newblock \bibinfo{journal}{\emph{arXiv preprint arXiv:1504.00325}} (\bibinfo{year}{2015}).
\newblock


\bibitem[Chu et~al\mbox{.}(2024)]%
        {chu2024qwen2}
\bibfield{author}{\bibinfo{person}{Y. Chu}, \bibinfo{person}{J. Xu}, \bibinfo{person}{Q. Yang}, \bibinfo{person}{H. Wei}, \bibinfo{person}{X. Wei}, \bibinfo{person}{Z. Guo}, \bibinfo{person}{Y. Leng}, \bibinfo{person}{Y. Lv}, \bibinfo{person}{J. He}, \bibinfo{person}{J. Lin}, {et~al\mbox{.}}} \bibinfo{year}{2024}\natexlab{}.
\newblock \showarticletitle{Qwen2-audio technical report}.
\newblock \bibinfo{journal}{\emph{arXiv preprint arXiv:2407.10759}} (\bibinfo{year}{2024}).
\newblock


\bibitem[Doh et~al\mbox{.}(2023)]%
        {doh2023lp}
\bibfield{author}{\bibinfo{person}{S. Doh}, \bibinfo{person}{K. Choi}, \bibinfo{person}{J. Lee}, {and} \bibinfo{person}{J. Nam}.} \bibinfo{year}{2023}\natexlab{}.
\newblock \showarticletitle{LP-MusicCaps: LLM-Based Pseudo Music Captioning}. In \bibinfo{booktitle}{\emph{ISMIR}}.
\newblock


\bibitem[Drossos et~al\mbox{.}(2020)]%
        {drossos2020clotho}
\bibfield{author}{\bibinfo{person}{Konstantinos Drossos}, \bibinfo{person}{Samuel Lipping}, {and} \bibinfo{person}{Tuomas Virtanen}.} \bibinfo{year}{2020}\natexlab{}.
\newblock \showarticletitle{Clotho: An audio captioning dataset}. In \bibinfo{booktitle}{\emph{ICASSP 2020-2020 IEEE International Conference on Acoustics, Speech and Signal Processing (ICASSP)}}. IEEE, \bibinfo{pages}{736--740}.
\newblock


\bibitem[Duan et~al\mbox{.}(2013)]%
        {6694316}
\bibfield{author}{\bibinfo{person}{Zhiyan Duan}, \bibinfo{person}{Haotian Fang}, \bibinfo{person}{Bo Li}, \bibinfo{person}{Khe~Chai Sim}, {and} \bibinfo{person}{Ye Wang}.} \bibinfo{year}{2013}\natexlab{}.
\newblock \showarticletitle{The NUS sung and spoken lyrics corpus: A quantitative comparison of singing and speech}. In \bibinfo{booktitle}{\emph{2013 Asia-Pacific Signal and Information Processing Association Annual Summit and Conference}}. \bibinfo{pages}{1--9}.
\newblock
\href{https://doi.org/10.1109/APSIPA.2013.6694316}{doi:\nolinkurl{10.1109/APSIPA.2013.6694316}}


\bibitem[Dubey et~al\mbox{.}(2024)]%
        {dubey2024llama}
\bibfield{author}{\bibinfo{person}{Abhimanyu Dubey}, \bibinfo{person}{Abhinav Jauhri}, \bibinfo{person}{Abhinav Pandey}, \bibinfo{person}{Abhishek Kadian}, \bibinfo{person}{Ahmad Al-Dahle}, \bibinfo{person}{Aiesha Letman}, \bibinfo{person}{Akhil Mathur}, \bibinfo{person}{Alan Schelten}, \bibinfo{person}{Amy Yang}, \bibinfo{person}{Angela Fan}, {et~al\mbox{.}}} \bibinfo{year}{2024}\natexlab{}.
\newblock \showarticletitle{The llama 3 herd of models}.
\newblock \bibinfo{journal}{\emph{arXiv preprint arXiv:2407.21783}} (\bibinfo{year}{2024}).
\newblock


\bibitem[Ferraro et~al\mbox{.}(2021)]%
        {ferraro2021melon}
\bibfield{author}{\bibinfo{person}{A. Ferraro}, \bibinfo{person}{Y. Kim}, \bibinfo{person}{S. Lee}, \bibinfo{person}{B. Kim}, \bibinfo{person}{N. Jo}, \bibinfo{person}{S. Lim}, \bibinfo{person}{S. Lim}, \bibinfo{person}{J. Jang}, \bibinfo{person}{S. Kim}, \bibinfo{person}{X. Serra}, {et~al\mbox{.}}} \bibinfo{year}{2021}\natexlab{}.
\newblock \showarticletitle{Melon playlist dataset: A public dataset for audio-based playlist generation and music tagging}. In \bibinfo{booktitle}{\emph{ICASSP}}.
\newblock


\bibitem[Guo et~al\mbox{.}(2023)]%
        {guo2023prompttts}
\bibfield{author}{\bibinfo{person}{Z. Guo}, \bibinfo{person}{Y. Leng}, \bibinfo{person}{Y. Wu}, \bibinfo{person}{S. Zhao}, {and} \bibinfo{person}{X. Tan}.} \bibinfo{year}{2023}\natexlab{}.
\newblock \showarticletitle{Prompttts: Controllable text-to-speech with text descriptions}. In \bibinfo{booktitle}{\emph{ICASSP}}.
\newblock


\bibitem[Hsu et~al\mbox{.}(2021)]%
        {hsu2021hubert}
\bibfield{author}{\bibinfo{person}{W.-N. Hsu}, \bibinfo{person}{B. Bolte}, \bibinfo{person}{Y.-H.~H. Tsai}, \bibinfo{person}{K. Lakhotia}, \bibinfo{person}{R. Salakhutdinov}, {and} \bibinfo{person}{A. Mohamed}.} \bibinfo{year}{2021}\natexlab{}.
\newblock \showarticletitle{Hubert: Self-supervised speech representation learning by masked prediction of hidden units}.
\newblock \bibinfo{journal}{\emph{IEEE/ACM TASLP}} (\bibinfo{year}{2021}).
\newblock


\bibitem[Huang et~al\mbox{.}(2021)]%
        {huang2021multi}
\bibfield{author}{\bibinfo{person}{Rongjie Huang}, \bibinfo{person}{Feiyang Chen}, \bibinfo{person}{Yi Ren}, \bibinfo{person}{Jinglin Liu}, \bibinfo{person}{Chenye Cui}, {and} \bibinfo{person}{Zhou Zhao}.} \bibinfo{year}{2021}\natexlab{}.
\newblock \showarticletitle{Multi-singer: Fast multi-singer singing voice vocoder with a large-scale corpus}. In \bibinfo{booktitle}{\emph{Proceedings of the 29th ACM International Conference on Multimedia}}. \bibinfo{pages}{3945--3954}.
\newblock


\bibitem[Hurst et~al\mbox{.}(2024)]%
        {hurst2024gpt}
\bibfield{author}{\bibinfo{person}{Aaron Hurst}, \bibinfo{person}{Adam Lerer}, \bibinfo{person}{Adam~P Goucher}, \bibinfo{person}{Adam Perelman}, \bibinfo{person}{Aditya Ramesh}, \bibinfo{person}{Aidan Clark}, \bibinfo{person}{AJ Ostrow}, \bibinfo{person}{Akila Welihinda}, \bibinfo{person}{Alan Hayes}, \bibinfo{person}{Alec Radford}, {et~al\mbox{.}}} \bibinfo{year}{2024}\natexlab{}.
\newblock \showarticletitle{Gpt-4o system card}.
\newblock \bibinfo{journal}{\emph{arXiv preprint arXiv:2410.21276}} (\bibinfo{year}{2024}).
\newblock


\bibitem[Kim et~al\mbox{.}(2019)]%
        {kim2019audiocaps}
\bibfield{author}{\bibinfo{person}{C.~D. Kim}, \bibinfo{person}{B. Kim}, \bibinfo{person}{H. Lee}, {and} \bibinfo{person}{G. Kim}.} \bibinfo{year}{2019}\natexlab{}.
\newblock \showarticletitle{Audiocaps: Generating captions for audios in the wild}. In \bibinfo{booktitle}{\emph{NAACL}}.
\newblock


\bibitem[Kim et~al\mbox{.}(2023)]%
        {kim2023prefix}
\bibfield{author}{\bibinfo{person}{M. Kim}, \bibinfo{person}{K. Sung-Bin}, {and} \bibinfo{person}{T.-H. Oh}.} \bibinfo{year}{2023}\natexlab{}.
\newblock \showarticletitle{Prefix tuning for automated audio captioning}. In \bibinfo{booktitle}{\emph{ICASSP}}.
\newblock


\bibitem[Kolchinski et~al\mbox{.}(2019)]%
        {kolchinski2019approximating}
\bibfield{author}{\bibinfo{person}{Y~Alex Kolchinski}, \bibinfo{person}{Sharon Zhou}, \bibinfo{person}{Shengjia Zhao}, \bibinfo{person}{Mitchell Gordon}, {and} \bibinfo{person}{Stefano Ermon}.} \bibinfo{year}{2019}\natexlab{}.
\newblock \showarticletitle{Approximating human judgment of generated image quality}.
\newblock \bibinfo{journal}{\emph{arXiv preprint arXiv:1912.12121}} (\bibinfo{year}{2019}).
\newblock


\bibitem[Leng et~al\mbox{.}(2024)]%
        {leng2024prompttts}
\bibfield{author}{\bibinfo{person}{Y. Leng}, \bibinfo{person}{Z. Guo}, \bibinfo{person}{K. Shen}, \bibinfo{person}{Z. Ju}, \bibinfo{person}{X. Tan}, \bibinfo{person}{E. Liu}, \bibinfo{person}{Y. Liu}, \bibinfo{person}{D. Yang}, \bibinfo{person}{l. zhang}, \bibinfo{person}{K. Song}, \bibinfo{person}{L. He}, \bibinfo{person}{X. Li}, \bibinfo{person}{s. zhao}, \bibinfo{person}{T. Qin}, {and} \bibinfo{person}{J. Bian}.} \bibinfo{year}{2024}\natexlab{}.
\newblock \showarticletitle{Prompt{TTS} 2: Describing and Generating Voices with Text Prompt}. In \bibinfo{booktitle}{\emph{ICLR}}.
\newblock
\urldef\tempurl%
\url{https://openreview.net/forum?id=NsCXDyv2Bn}
\showURL{%
\tempurl}


\bibitem[LI et~al\mbox{.}(2024)]%
        {li2024mert}
\bibfield{author}{\bibinfo{person}{Y. LI}, \bibinfo{person}{R. Yuan}, \bibinfo{person}{G. Zhang}, \bibinfo{person}{Y. Ma}, \bibinfo{person}{X. Chen}, \bibinfo{person}{H. Yin}, \bibinfo{person}{C. Xiao}, \bibinfo{person}{C. Lin}, \bibinfo{person}{A. Ragni}, \bibinfo{person}{E. Benetos}, \bibinfo{person}{N. Gyenge}, \bibinfo{person}{R. Dannenberg}, \bibinfo{person}{R. Liu}, \bibinfo{person}{W. Chen}, \bibinfo{person}{G. Xia}, \bibinfo{person}{Y. Shi}, \bibinfo{person}{W. Huang}, \bibinfo{person}{Z. Wang}, \bibinfo{person}{Y. Guo}, {and} \bibinfo{person}{J. Fu}.} \bibinfo{year}{2024}\natexlab{}.
\newblock \showarticletitle{{MERT}: Acoustic Music Understanding Model with Large-Scale Self-supervised Training}. In \bibinfo{booktitle}{\emph{ICLR}}.
\newblock
\urldef\tempurl%
\url{https://openreview.net/forum?id=w3YZ9MSlBu}
\showURL{%
\tempurl}


\bibitem[Lin(2004)]%
        {lin-2004-rouge}
\bibfield{author}{\bibinfo{person}{C.-Y. Lin}.} \bibinfo{year}{2004}\natexlab{}.
\newblock \showarticletitle{{ROUGE}: A Package for Automatic Evaluation of Summaries}. In \bibinfo{booktitle}{\emph{Text Summarization Branches Out}}.
\newblock
\urldef\tempurl%
\url{https://aclanthology.org/W04-1013}
\showURL{%
\tempurl}


\bibitem[Liu et~al\mbox{.}(2017)]%
        {liu2017improved}
\bibfield{author}{\bibinfo{person}{S. Liu}, \bibinfo{person}{Z. Zhu}, \bibinfo{person}{N. Ye}, \bibinfo{person}{S. Guadarrama}, {and} \bibinfo{person}{K. Murphy}.} \bibinfo{year}{2017}\natexlab{}.
\newblock \showarticletitle{Improved image captioning via policy gradient optimization of spider}. In \bibinfo{booktitle}{\emph{ICCV}}.
\newblock


\bibitem[Luo and Yu(2023)]%
        {luo2023music}
\bibfield{author}{\bibinfo{person}{Yi Luo} {and} \bibinfo{person}{Jianwei Yu}.} \bibinfo{year}{2023}\natexlab{}.
\newblock \showarticletitle{Music source separation with band-split RNN}.
\newblock \bibinfo{journal}{\emph{IEEE/ACM Transactions on Audio, Speech, and Language Processing}}  \bibinfo{volume}{31} (\bibinfo{year}{2023}), \bibinfo{pages}{1893--1901}.
\newblock


\bibitem[Luo et~al\mbox{.}(2020)]%
        {luo2020singing}
\bibfield{author}{\bibinfo{person}{Y.-J. Luo}, \bibinfo{person}{C.-C. Hsu}, \bibinfo{person}{K. Agres}, {and} \bibinfo{person}{D. Herremans}.} \bibinfo{year}{2020}\natexlab{}.
\newblock \showarticletitle{Singing voice conversion with disentangled representations of singer and vocal technique using variational autoencoders}. In \bibinfo{booktitle}{\emph{ICASSP}}.
\newblock


\bibitem[Papineni et~al\mbox{.}(2002)]%
        {10.3115/1073083.1073135}
\bibfield{author}{\bibinfo{person}{K. Papineni}, \bibinfo{person}{S. Roukos}, \bibinfo{person}{T. Ward}, {and} \bibinfo{person}{W.-J. Zhu}.} \bibinfo{year}{2002}\natexlab{}.
\newblock \showarticletitle{BLEU: A Method for Automatic Evaluation of Machine Translation}. In \bibinfo{booktitle}{\emph{ACL}}.
\newblock
\href{https://doi.org/10.3115/1073083.1073135}{doi:\nolinkurl{10.3115/1073083.1073135}}


\bibitem[Raffel et~al\mbox{.}(2020)]%
        {raffel2020exploring}
\bibfield{author}{\bibinfo{person}{Colin Raffel}, \bibinfo{person}{Noam Shazeer}, \bibinfo{person}{Adam Roberts}, \bibinfo{person}{Katherine Lee}, \bibinfo{person}{Sharan Narang}, \bibinfo{person}{Michael Matena}, \bibinfo{person}{Yanqi Zhou}, \bibinfo{person}{Wei Li}, {and} \bibinfo{person}{Peter~J Liu}.} \bibinfo{year}{2020}\natexlab{}.
\newblock \showarticletitle{Exploring the limits of transfer learning with a unified text-to-text transformer}.
\newblock \bibinfo{journal}{\emph{Journal of machine learning research}} \bibinfo{volume}{21}, \bibinfo{number}{140} (\bibinfo{year}{2020}), \bibinfo{pages}{1--67}.
\newblock


\bibitem[Reimers and Gurevych(2019)]%
        {reimers-gurevych-2019-sentence}
\bibfield{author}{\bibinfo{person}{N. Reimers} {and} \bibinfo{person}{I. Gurevych}.} \bibinfo{year}{2019}\natexlab{}.
\newblock \showarticletitle{Sentence-{BERT}: Sentence Embeddings using {S}iamese {BERT}-Networks}. In \bibinfo{booktitle}{\emph{EMNLP-IJCNLP}}.
\newblock
\href{https://doi.org/10.18653/v1/D19-1410}{doi:\nolinkurl{10.18653/v1/D19-1410}}


\bibitem[Rohrbach et~al\mbox{.}(2017)]%
        {rohrbach2017movie}
\bibfield{author}{\bibinfo{person}{Anna Rohrbach}, \bibinfo{person}{Atousa Torabi}, \bibinfo{person}{Marcus Rohrbach}, \bibinfo{person}{Niket Tandon}, \bibinfo{person}{Christopher Pal}, \bibinfo{person}{Hugo Larochelle}, \bibinfo{person}{Aaron Courville}, {and} \bibinfo{person}{Bernt Schiele}.} \bibinfo{year}{2017}\natexlab{}.
\newblock \showarticletitle{Movie description}.
\newblock \bibinfo{journal}{\emph{International Journal of Computer Vision}}  \bibinfo{volume}{123} (\bibinfo{year}{2017}), \bibinfo{pages}{94--120}.
\newblock


\bibitem[Rouard et~al\mbox{.}(2023)]%
        {rouard2023hybrid}
\bibfield{author}{\bibinfo{person}{S. Rouard}, \bibinfo{person}{F. Massa}, {and} \bibinfo{person}{A. D{\'e}fossez}.} \bibinfo{year}{2023}\natexlab{}.
\newblock \showarticletitle{Hybrid transformers for music source separation}. In \bibinfo{booktitle}{\emph{ICASSP}}.
\newblock


\bibitem[Sharma et~al\mbox{.}(2018)]%
        {sharma-etal-2018-conceptual}
\bibfield{author}{\bibinfo{person}{Piyush Sharma}, \bibinfo{person}{Nan Ding}, \bibinfo{person}{Sebastian Goodman}, {and} \bibinfo{person}{Radu Soricut}.} \bibinfo{year}{2018}\natexlab{}.
\newblock \showarticletitle{Conceptual Captions: A Cleaned, Hypernymed, Image Alt-text Dataset For Automatic Image Captioning}. In \bibinfo{booktitle}{\emph{Proceedings of the 56th Annual Meeting of the Association for Computational Linguistics (Volume 1: Long Papers)}}, \bibfield{editor}{\bibinfo{person}{Iryna Gurevych} {and} \bibinfo{person}{Yusuke Miyao}} (Eds.). \bibinfo{publisher}{Association for Computational Linguistics}, \bibinfo{address}{Melbourne, Australia}, \bibinfo{pages}{2556--2565}.
\newblock
\href{https://doi.org/10.18653/v1/P18-1238}{doi:\nolinkurl{10.18653/v1/P18-1238}}


\bibitem[Vedantam et~al\mbox{.}(2015)]%
        {vedantam2015cider}
\bibfield{author}{\bibinfo{person}{R. Vedantam}, \bibinfo{person}{C. Lawrence~Zitnick}, {and} \bibinfo{person}{D. Parikh}.} \bibinfo{year}{2015}\natexlab{}.
\newblock \showarticletitle{Cider: Consensus-based image description evaluation}. In \bibinfo{booktitle}{\emph{CVPR}}.
\newblock


\bibitem[Wang et~al\mbox{.}(2024)]%
        {wang2024prompt}
\bibfield{author}{\bibinfo{person}{Y. Wang}, \bibinfo{person}{R. Hu}, \bibinfo{person}{R. Huang}, \bibinfo{person}{Z. Hong}, \bibinfo{person}{R. Li}, \bibinfo{person}{W. Liu}, \bibinfo{person}{F. You}, \bibinfo{person}{T. Jin}, {and} \bibinfo{person}{Z. Zhao}.} \bibinfo{year}{2024}\natexlab{}.
\newblock \showarticletitle{Prompt-Singer: Controllable Singing-Voice-Synthesis with Natural Language Prompt}. In \bibinfo{booktitle}{\emph{NAACL}}.
\newblock


\bibitem[Wang et~al\mbox{.}(2018)]%
        {wang2018style}
\bibfield{author}{\bibinfo{person}{Y. Wang}, \bibinfo{person}{D. Stanton}, \bibinfo{person}{Y. Zhang}, \bibinfo{person}{R.-S. Ryan}, \bibinfo{person}{E. Battenberg}, \bibinfo{person}{J. Shor}, \bibinfo{person}{Y. Xiao}, \bibinfo{person}{Y. Jia}, \bibinfo{person}{F. Ren}, {and} \bibinfo{person}{R.~A Saurous}.} \bibinfo{year}{2018}\natexlab{}.
\newblock \showarticletitle{Style tokens: Unsupervised style modeling, control and transfer in end-to-end speech synthesis}. In \bibinfo{booktitle}{\emph{ICML}}.
\newblock


\bibitem[Wang et~al\mbox{.}(2022)]%
        {wang2022opencpop}
\bibfield{author}{\bibinfo{person}{Yu Wang}, \bibinfo{person}{Xinsheng Wang}, \bibinfo{person}{Pengcheng Zhu}, \bibinfo{person}{Jie Wu}, \bibinfo{person}{Hanzhao Li}, \bibinfo{person}{Heyang Xue}, \bibinfo{person}{Yongmao Zhang}, \bibinfo{person}{Lei Xie}, {and} \bibinfo{person}{Mengxiao Bi}.} \bibinfo{year}{2022}\natexlab{}.
\newblock \showarticletitle{Opencpop: A high-quality open source chinese popular song corpus for singing voice synthesis}.
\newblock \bibinfo{journal}{\emph{arXiv preprint arXiv:2201.07429}} (\bibinfo{year}{2022}).
\newblock


\bibitem[Xu et~al\mbox{.}(2016)]%
        {7780940}
\bibfield{author}{\bibinfo{person}{Jun Xu}, \bibinfo{person}{Tao Mei}, \bibinfo{person}{Ting Yao}, {and} \bibinfo{person}{Yong Rui}.} \bibinfo{year}{2016}\natexlab{}.
\newblock \showarticletitle{MSR-VTT: A Large Video Description Dataset for Bridging Video and Language}. In \bibinfo{booktitle}{\emph{2016 IEEE Conference on Computer Vision and Pattern Recognition (CVPR)}}. \bibinfo{pages}{5288--5296}.
\newblock
\href{https://doi.org/10.1109/CVPR.2016.571}{doi:\nolinkurl{10.1109/CVPR.2016.571}}


\bibitem[Yamauchi et~al\mbox{.}(2024)]%
        {yamauchi2024stylecap}
\bibfield{author}{\bibinfo{person}{K. Yamauchi}, \bibinfo{person}{Y. Ijima}, {and} \bibinfo{person}{Y. Saito}.} \bibinfo{year}{2024}\natexlab{}.
\newblock \showarticletitle{StyleCap: Automatic Speaking-Style Captioning from Speech Based on Speech and Language Self-supervised Learning Models}. In \bibinfo{booktitle}{\emph{ICASSP}}.
\newblock


\bibitem[Yang et~al\mbox{.}(2024c)]%
        {yang2024qwen2}
\bibfield{author}{\bibinfo{person}{A. Yang}, \bibinfo{person}{B. Yang}, \bibinfo{person}{B. Hui}, \bibinfo{person}{B. Zheng}, \bibinfo{person}{B. Yu}, \bibinfo{person}{C. Zhou}, \bibinfo{person}{C. Li}, \bibinfo{person}{C. Li}, \bibinfo{person}{D. Liu}, \bibinfo{person}{F. Huang}, {et~al\mbox{.}}} \bibinfo{year}{2024}\natexlab{c}.
\newblock \showarticletitle{Qwen2 technical report}.
\newblock \bibinfo{journal}{\emph{arXiv preprint arXiv:2407.10671}} (\bibinfo{year}{2024}).
\newblock


\bibitem[Yang et~al\mbox{.}(2024a)]%
        {yang2024instructtts}
\bibfield{author}{\bibinfo{person}{D. Yang}, \bibinfo{person}{S. Liu}, \bibinfo{person}{R. Huang}, \bibinfo{person}{C. Weng}, {and} \bibinfo{person}{H. Meng}.} \bibinfo{year}{2024}\natexlab{a}.
\newblock \showarticletitle{Instructtts: Modelling expressive tts in discrete latent space with natural language style prompt}.
\newblock \bibinfo{journal}{\emph{IEEE/ACM TASLP}} (\bibinfo{year}{2024}).
\newblock


\bibitem[Yang et~al\mbox{.}(2024b)]%
        {yang2024air}
\bibfield{author}{\bibinfo{person}{Qian Yang}, \bibinfo{person}{Jin Xu}, \bibinfo{person}{Wenrui Liu}, \bibinfo{person}{Yunfei Chu}, \bibinfo{person}{Ziyue Jiang}, \bibinfo{person}{Xiaohuan Zhou}, \bibinfo{person}{Yichong Leng}, \bibinfo{person}{Yuanjun Lv}, \bibinfo{person}{Zhou Zhao}, \bibinfo{person}{Chang Zhou}, {et~al\mbox{.}}} \bibinfo{year}{2024}\natexlab{b}.
\newblock \showarticletitle{Air-bench: Benchmarking large audio-language models via generative comprehension}.
\newblock \bibinfo{journal}{\emph{arXiv preprint arXiv:2402.07729}} (\bibinfo{year}{2024}).
\newblock


\bibitem[Yao et~al\mbox{.}(2024)]%
        {yao2024promptvc}
\bibfield{author}{\bibinfo{person}{Jixun Yao}, \bibinfo{person}{Yuguang Yang}, \bibinfo{person}{Yi Lei}, \bibinfo{person}{Ziqian Ning}, \bibinfo{person}{Yanni Hu}, \bibinfo{person}{Yu Pan}, \bibinfo{person}{Jingjing Yin}, \bibinfo{person}{Hongbin Zhou}, \bibinfo{person}{Heng Lu}, {and} \bibinfo{person}{Lei Xie}.} \bibinfo{year}{2024}\natexlab{}.
\newblock \showarticletitle{Promptvc: Flexible stylistic voice conversion in latent space driven by natural language prompts}. In \bibinfo{booktitle}{\emph{ICASSP 2024-2024 IEEE International Conference on Acoustics, Speech and Signal Processing (ICASSP)}}. IEEE, \bibinfo{pages}{10571--10575}.
\newblock


\bibitem[Young et~al\mbox{.}(2014)]%
        {young-etal-2014-image}
\bibfield{author}{\bibinfo{person}{Peter Young}, \bibinfo{person}{Alice Lai}, \bibinfo{person}{Micah Hodosh}, {and} \bibinfo{person}{Julia Hockenmaier}.} \bibinfo{year}{2014}\natexlab{}.
\newblock \showarticletitle{From image descriptions to visual denotations: New similarity metrics for semantic inference over event descriptions}.
\newblock \bibinfo{journal}{\emph{Transactions of the Association for Computational Linguistics}}  \bibinfo{volume}{2} (\bibinfo{year}{2014}), \bibinfo{pages}{67--78}.
\newblock
\href{https://doi.org/10.1162/tacl_a_00166}{doi:\nolinkurl{10.1162/tacl_a_00166}}


\bibitem[Zhang et~al\mbox{.}(2022)]%
        {NEURIPS2022_2de60892}
\bibfield{author}{\bibinfo{person}{Lichao Zhang}, \bibinfo{person}{Ruiqi Li}, \bibinfo{person}{Shoutong Wang}, \bibinfo{person}{Liqun Deng}, \bibinfo{person}{Jinglin Liu}, \bibinfo{person}{Yi Ren}, \bibinfo{person}{Jinzheng He}, \bibinfo{person}{Rongjie Huang}, \bibinfo{person}{Jieming Zhu}, \bibinfo{person}{Xiao Chen}, {and} \bibinfo{person}{Zhou Zhao}.} \bibinfo{year}{2022}\natexlab{}.
\newblock \showarticletitle{M4Singer: A Multi-Style, Multi-Singer and Musical Score Provided Mandarin Singing Corpus}. In \bibinfo{booktitle}{\emph{Advances in Neural Information Processing Systems}}, \bibfield{editor}{\bibinfo{person}{S.~Koyejo}, \bibinfo{person}{S.~Mohamed}, \bibinfo{person}{A.~Agarwal}, \bibinfo{person}{D.~Belgrave}, \bibinfo{person}{K.~Cho}, {and} \bibinfo{person}{A.~Oh}} (Eds.), Vol.~\bibinfo{volume}{35}. \bibinfo{publisher}{Curran Associates, Inc.}, \bibinfo{pages}{6914--6926}.
\newblock
\urldef\tempurl%
\url{https://proceedings.neurips.cc/paper_files/paper/2022/file/2de60892dd329683ec21877a4e7c3091-Paper-Datasets_and_Benchmarks.pdf}
\showURL{%
\tempurl}


\bibitem[Zhang et~al\mbox{.}(2024)]%
        {zhang2024stylesinger}
\bibfield{author}{\bibinfo{person}{Y. Zhang}, \bibinfo{person}{R. Huang}, \bibinfo{person}{R. Li}, \bibinfo{person}{J. He}, \bibinfo{person}{Y. Xia}, \bibinfo{person}{F. Chen}, \bibinfo{person}{X. Duan}, \bibinfo{person}{B. Huai}, {and} \bibinfo{person}{Z. Zhao}.} \bibinfo{year}{2024}\natexlab{}.
\newblock \showarticletitle{StyleSinger: Style Transfer for Out-of-Domain Singing Voice Synthesis}. In \bibinfo{booktitle}{\emph{AAAI}}.
\newblock


\bibitem[Zhang et~al\mbox{.}(2023)]%
        {zhang2023promptspeaker}
\bibfield{author}{\bibinfo{person}{Yongmao Zhang}, \bibinfo{person}{Guanghou Liu}, \bibinfo{person}{Yi Lei}, \bibinfo{person}{Yunlin Chen}, \bibinfo{person}{Hao Yin}, \bibinfo{person}{Lei Xie}, {and} \bibinfo{person}{Zhifei Li}.} \bibinfo{year}{2023}\natexlab{}.
\newblock \showarticletitle{Promptspeaker: Speaker generation based on text descriptions}. In \bibinfo{booktitle}{\emph{2023 IEEE Automatic Speech Recognition and Understanding Workshop (ASRU)}}. IEEE, \bibinfo{pages}{1--7}.
\newblock


\bibitem[Zhou et~al\mbox{.}(2022)]%
        {zhou2022can}
\bibfield{author}{\bibinfo{person}{Z. Zhou}, \bibinfo{person}{Z. Zhang}, \bibinfo{person}{X. Xu}, \bibinfo{person}{Z. Xie}, \bibinfo{person}{M. Wu}, {and} \bibinfo{person}{K.~Q Zhu}.} \bibinfo{year}{2022}\natexlab{}.
\newblock \showarticletitle{Can audio captions be evaluated with image caption metrics?}. In \bibinfo{booktitle}{\emph{ICASSP}}.
\newblock


\bibitem[Zhu et~al\mbox{.}(2024)]%
        {zhu2024unistyle}
\bibfield{author}{\bibinfo{person}{X. Zhu}, \bibinfo{person}{W. Tian}, \bibinfo{person}{X. Wang}, \bibinfo{person}{L. He}, \bibinfo{person}{Y. Xiao}, \bibinfo{person}{X. Wang}, \bibinfo{person}{X. Tan}, \bibinfo{person}{L. Xie}, {et~al\mbox{.}}} \bibinfo{year}{2024}\natexlab{}.
\newblock \showarticletitle{UniStyle: Unified Style Modeling for Speaking Style Captioning and Stylistic Speech Synthesis}. In \bibinfo{booktitle}{\emph{ACM MM}}.
\newblock


\end{thebibliography}

\end{document}